\documentclass[letterpaper]{article} % DO NOT CHANGE THIS
\usepackage{aaai24}  % DO NOT CHANGE THIS
\usepackage{times}  % DO NOT CHANGE THIS
\usepackage{helvet}  % DO NOT CHANGE THIS
\usepackage{courier}  % DO NOT CHANGE THIS
\usepackage[hyphens]{url}  % DO NOT CHANGE THIS
\usepackage{graphicx} % DO NOT CHANGE THIS
\usepackage{natbib}  % DO NOT CHANGE THIS AND DO NOT ADD ANY OPTIONS TO IT
\usepackage{caption} % DO NOT CHANGE THIS AND DO NOT ADD ANY OPTIONS TO IT
\usepackage{algorithm}
\usepackage{algorithmic}

\usepackage{float}
\usepackage{subfig}
\usepackage{booktabs} % for professional tables
\usepackage{amsmath}
\usepackage{amsthm}
\usepackage{amsfonts}       % blackboard math symbols
\usepackage{nicefrac}       % compact symbols for 1/2, etc.
\usepackage{xspace}
\usepackage{xcolor}
\usepackage{backnaur}
\usepackage{multirow}
\usepackage{makecell}
\usepackage{appendix}
\usepackage{enumitem}
\usepackage{circledsteps}
\usepackage{color}
\usepackage{colortbl}
\usepackage{graphicx}
\usepackage{fixltx2e}
\usepackage{amsmath}
\usepackage{caption}  
\usepackage{listings}
\usepackage{graphicx}

\newcommand{\reffig}[1]{Figure~\ref{#1}}
\newcommand{\refsec}[1]{\S\ref{#1}} % \textsection
\newcommand{\reftab}[1]{Table~\ref{#1}}
\def\eg{\textit{e.g.}\xspace}
\def\Eg{\textit{E.g.}\xspace}

\makeatletter
\newcommand{\printfnsymbol}[1]{%
  \textsuperscript{\@fnsymbol{#1}}%
} 
\makeatother

\usepackage{newfloat}
\usepackage{listings}
\DeclareCaptionStyle{ruled}{labelfont=normalfont,labelsep=colon,strut=off} % DO NOT CHANGE THIS
\floatstyle{ruled}
\newfloat{listing}{tb}{lst}{}
\floatname{listing}{Listing}
\title{Text2Analysis: A Benchmark of Table Question Answering with Advanced~Data~Analysis and Unclear~Queries}
\author{
    Xinyi He\textsuperscript{\rm 1}\thanks{\indent The contributions by Xinyi He and Xinrun Xu have been conducted and completed during their internships at Microsoft.},
    Mengyu Zhou\textsuperscript{\rm 2}\thanks{\indent Corresponding author.},
    Xinrun Xu\textsuperscript{\rm 3}\printfnsymbol{1},
    Xiaojun Ma\textsuperscript{\rm 2}, 
    Rui Ding\textsuperscript{\rm 2}, 
    Lun Du\textsuperscript{\rm 2}, \\
    Yan Gao\textsuperscript{\rm 2},
    Ran Jia\textsuperscript{\rm 2}, 
    Xu Chen\textsuperscript{\rm 2},
    Shi Han\textsuperscript{\rm 2},
    Zejian Yuan\textsuperscript{\rm 1},
    Dongmei Zhang\textsuperscript{\rm 2}
}
\affiliations {
    \textsuperscript{\rm 1} Xi'an Jiaotong University, 
    \textsuperscript{\rm 2} Microsoft,
    \textsuperscript{\rm 3} Institute of Software Chinese Academy of Science\\
    
    hxyhxy@stu.xjtu.edu.cn,
    xuxinrun20@mails.ucas.ac.cn,
    yuan.ze.jian@xjtu.edu.cn,\\
    \{mezho, xiaojunma, juding, lun.du, gaoya, raji, xu.chen, shihan,dongmeiz\}@microsoft.com
}

\usepackage{bibentry}
\begin{document}

\maketitle

% \ensheng{A suitable title is needed.}
\begin{abstract}

Tabular data analysis is crucial in various fields, and large language models show promise in this area. However, current research mostly focuses on rudimentary tasks like Text2SQL and TableQA, neglecting advanced analysis like forecasting and chart generation.
To address this gap, we developed the Text2Analysis benchmark, incorporating advanced analysis tasks that go beyond the SQL-compatible operations and require more in-depth analysis. We also develop five innovative and effective annotation methods, harnessing the capabilities of large language models to enhance data quality and quantity. Additionally, we include unclear queries that resemble real-world user questions to test how well models can understand and tackle such challenges. Finally, we collect 2249 query-result pairs with 347 tables. We evaluate five state-of-the-art models using three different metrics and the results show that our benchmark presents introduces considerable challenge  in the field of tabular data analysis,
% ~\ensheng{The conclusion is not good, clear and surprising. Should revise}
paving the way for more advanced research opportunities.

\end{abstract}

\section{Introduction}

Tabular data analysis plays a crucial role in various fields, and automated data analysis has the potential to enhance people's work efficiency significantly~\cite{delen2018research}. The emergence of large language models has shown promising capabilities to accelerate tabular data analysis~\cite{chen-2023-large,ye2023large, ma2023demonstration, Jiang-StructGPT-2022}. Understanding the analytical abilities of these models, identifying the analysis processes they can replace, and determining the analysis steps they can assist with have become pressing questions in the field.

\begin{figure}[htb]
    \centering
    \includegraphics[width=1\linewidth]{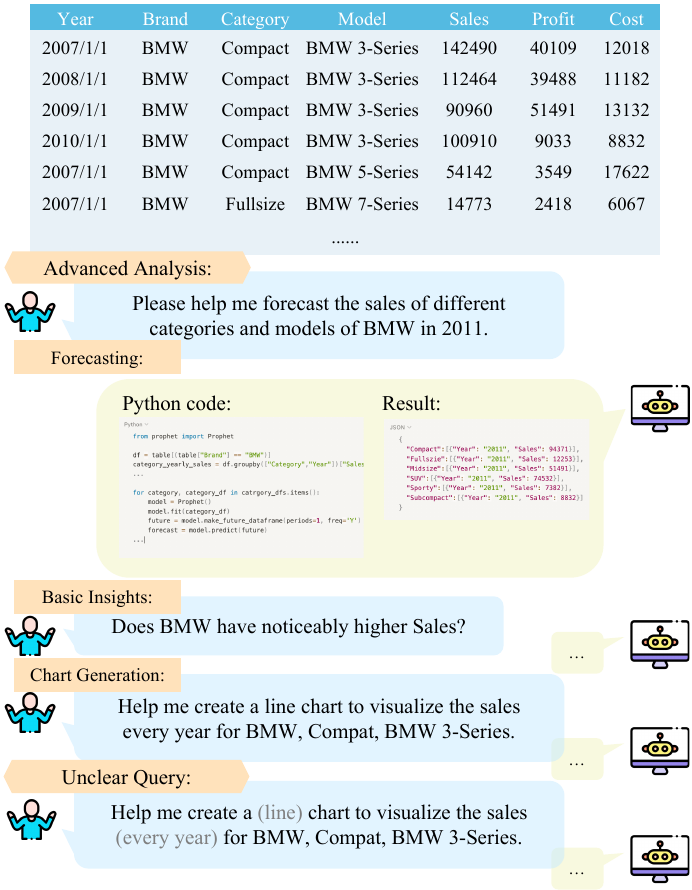}
    \caption{Examples of Text2Analysis Benchmark.}
    \label{fig:example}
    \vspace{-4mm}
\end{figure}

Existing research on tabular data analysis has limited coverage of data analysis. As shown in \reffig{fig:ana_task}, data analysis can be divided into descriptive, diagnostic, predictive, and prescriptive analytics~\cite{business_analytics}. The existing Text2SQL and TableQA datasets~\cite{dong2016nl2sql,km2021survey} focus primarily on rudimentary operations that are part of descriptive analytics and can be mostly solved by SQL and OLAP operations. They pay limited attention to \textbf{advanced analysis} (see \refsec{sec:ana_tax}) that require advanced operations and visualizations beyond rudimentary operations, such as calculating insights, forecasting, and chart generation (see examples in \reffig{fig:example}).

In the real world, many user queries are often described in unclear ways~\cite{wang-etal-2023-know}. When solving advanced or complex data analysis tasks with a large set of available tools and APIs, it is hardly the case that a user could write clear instructions with complete intent and parameters. 
As we will discuss in \refsec{sec:unclear_tax}, the most common \textbf{``unclear query''} type is missing parameters for analysis tasks. 
\Eg, the query ``Help me create a chart to visualize the sales for BMW, Compat, BMW 3-Serie'' does not explicitly specify the chart type to be drawn or the field to be mapped to the x axis. %This presents a significant challenge for models to parse and address these queries effectively.
%Thus, we also collect our dataset with ``\textbf{unclear queries}'' as described in \refsec{sec:unclear_tax}. 
Responding accurately to these queries not only demands the semantic parsing abilities of large language models but also requires them to possess strong data analysis capabilities to 
% discern the intent behind the query and 
recommend intent beyond the query.

In this paper, we propose the \textbf{Text2Analysis} benchmark which expands beyond rudimentary operations and clear instructions. The benchmark incorporates unclear queries that involve advanced data analysis. %These tasks can be combined and interconnected to form complex analyses. 
%To correctly solve these tasks, the problem has been formulated. The input is a table and a query, and the output is Python code and the corresponding result.
Similar to Text2SQL datasets, in Text2Analysis the input is the (table, query) pair, and the output is the (code, result) pair. The ground-truth code only leverages a set of chosen data analysis APIs / operations from public and customized Python libraries such as Pandas, Prophet and Matplotlib.

% Text2Analysis dataset contains tables, queries, corresponding Python code, and results. 
% We chose to use Python code because advanced analysis surpasses the language framework of SQL, which was originally employed in Text2SQL. Consequently, we opted for the more versatile and feature-rich Python to address these complex analytical tasks.

Collecting the dataset is a difficult task because each sample in the Text2Aanlysis dataset simultaneously contains a table, query, Python code, and result. It requires annotators with related expert backgrounds and would consume a lot of time. To accelerate the annotation process and increase the volume of annotated data, we have developed five innovative and reliable annotation methods. Those methods make full use of large language models to perform forward annotations, expansion with new tables, and expansion unclear queries. Meanwhile,
some methods also collect data from the output, such as reverse generation from codes or results. 
% These methods include Forward Annotation, Reverse Generation from Codes, Reverse Generation from Results, Expansion Unclear Queries, and Expansion Using Existing Data. 
We collect 2249 (query, code, result) pairs with 347 tables. To ensure annotation quality, iterative annotation and human evaluation are employed. Their results and dataset distribution indicate that Text2Analysis has a diverse, high-quality data analysis dataset.

% We chose to use Python code because advanced analysis surpasses the language framework of SQL, which was originally employed in Text2SQL. Consequently, we opted for the more versatile and feature-rich Python to address these complex analytical tasks.

Due to the numerous tasks involved in the problem and the outputs consisting of both code and results, evaluating the generated solutions with appropriate metrics poses a challenge. We have selected three metrics to evaluate from different perspectives: executable code ratio, pass rate, and regression metrics. 
The executable code ratio evaluates the model's ability to generate executable code. Pass rate evaluates the correctness of the generated code. Regression scores measure the predicting capability of the chosen model within the generated code.

Furthermore, we provide an evaluation of five current state-of-the-art models, including GPT family models, code generation models, and tabular models.
We evaluate their performance in handling advanced analysis and unclear queries. 
% Additionally, we conduct a detailed error analysis to gain insights into the challenges faced by these models in advanced analysis tasks. \TODO{Add insight}
Our experiment indicates that large language models exhibit robust parsing and code generation aptitudes for data analysis in the context of clear queries. However, they grapple with complex libraries and unclear queries. To augment their efficacy, future research can concentrate on bolstering the capacity to recommend fields for sophisticated analyses and tackling complex operations such as operations with complex parameter input and model training.

% The objective of this paper is to provide a comprehensive understanding of the analytical capabilities of large language models, specifically in advanced data analysis tasks. By addressing the existing research gap and conducting an in-depth analysis, we aim to shed light on the limitations and potential areas of improvement in utilizing large language models for advanced analysis. The outcomes of this study will contribute to the broader field of automated data analysis and facilitate the development of more effective models for enhanced analytical workflows.

In summary, our main contributions are:
\begin{itemize}
    \item We create the Text2Analysis benchmark which includes advanced analysis tasks and unclear queries that were rarely addressed in previous research work. %We definite the taxonomy of advanced analysis and unclear queries. 
    The dataset and code will be open-sourced on https://github.com/microsoft/Text2Analysis.
    \item We propose five innovative and reliable annotation methods for the construction of NL2Code datasets. They utilize large language models to accelerate the annotation process and increase the volume of annotation.
    \item The performance of the baseline models is systematically evaluated against our Text2Analysis benchmark. Our experiments show the challenges to be solved in the future to satisfy real-world table analysis needs.
\end{itemize}

\section{Problem Definition}
\label{sec:problem}

We introduce the Text2Analysis problem as follows: $(table,\ query) \rightarrow (code,\ result)$. The input consists of a $table$ and a user $query$. The output consists of Python $code$ snippet(s) and the corresponding $result$(s). A $table$ has n fields $T = (f_1, ..., f_n)$, and each field consists of a field header and field values.  A $query$ is related to data analysis, particularly focusing on advanced analysis (\refsec{sec:ana_tax}) that addresses the shortcomings of existing work and presents a greater and more difficult challenge for models. Additionally, it includes unclear queries (\refsec{sec:unclear_tax}), which are often found in real-world user scenarios and can more effectively evaluate the model's analytical capabilities.

\subsection{Analysis Operations and Tasks}
\label{sec:ana_tax}

\begin{figure}
    \centering
    \includegraphics[width=1\linewidth]{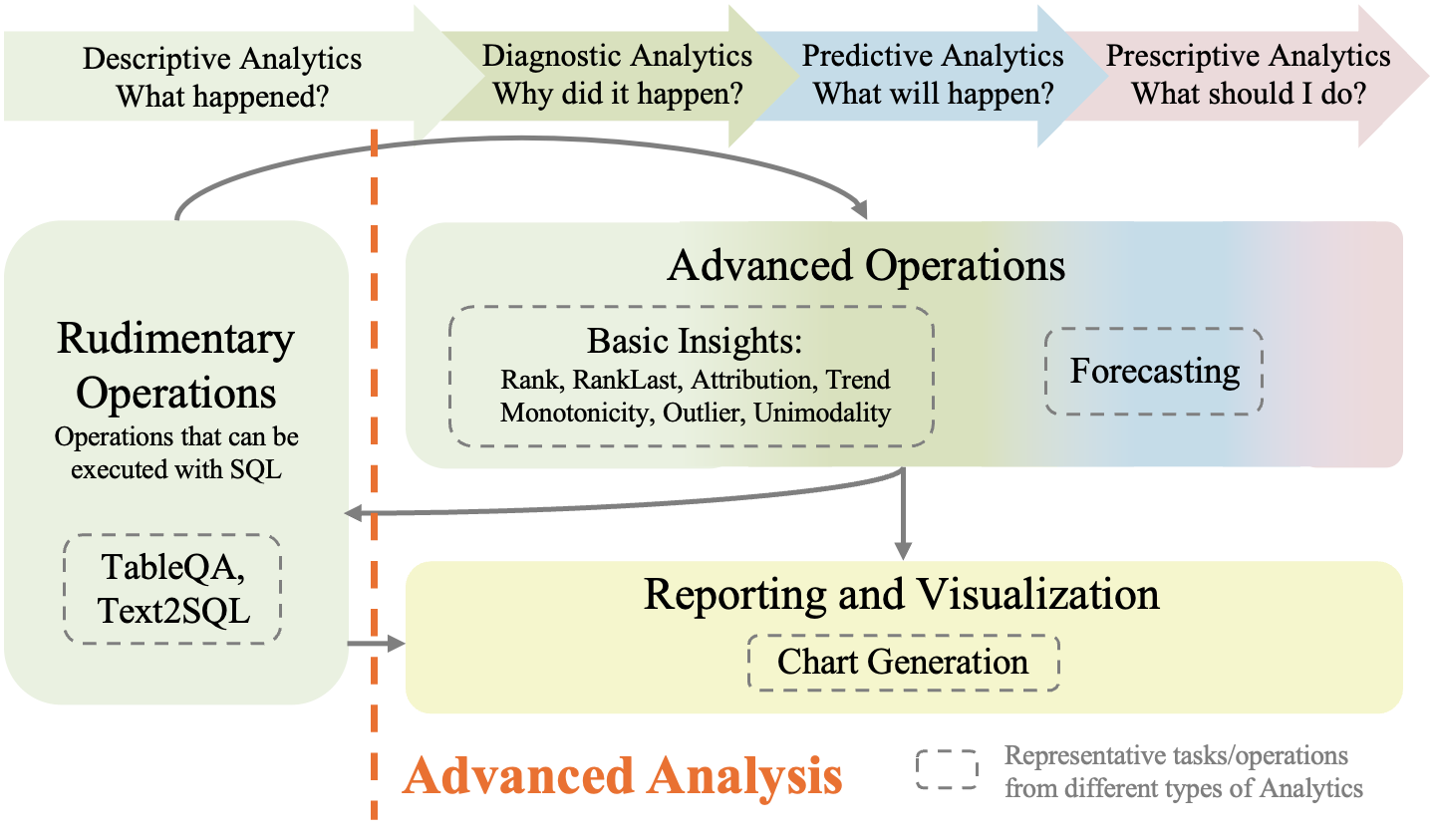}
    \caption{Advanced Analysis consists of Advanced Operations and Visualizations that are not covered by Rudimentary Operations across descriptive, diagnostic, predictive, and prescriptive analytics.}
    \label{fig:ana_task}
    \vspace{-4mm}
\end{figure}

Text2Analysis expands the data analysis dataset to advanced analysis tasks. As shown in \reffig{fig:ana_task}, Data analysis can be divided into descriptive (what happened?), diagnostic (why did it happen?), predictive(what will happen), and prescriptive analytics (what should I do?)~\cite{business_analytics}. And reporting and visualization may follow each type of analytics. Existing research on table-based data analysis tasks, such as TableQA and Text2SQL~\cite{dong2016nl2sql,km2021survey}, has focused mainly on part of  descriptive analytics that can be solved by SQL. They pay insufficient attention to advanced analysis that are beyond the rudimentary operations and require more in-depth analysis.  

The advanced analysis portion of Text2Analysis selects
% ~\ensheng{create or curate?} 
representative tasks from each type of analytics to form the dataset. From descriptive and diagnostic analytics, basic insights are chosen. From predictive analytics, forecasting is selected. And from reporting and visualization, chart generation is chosen. 
A more detailed introduction to each task will be provided after the following paragraph.

Advanced analysis, along with rudimentary operations, form the Text2Analysis dataset.  They can be combined to form a complex analysis. rudimentary operations and advanced operations (tasks in advanced analysis that output data such as tables and values, that is, tasks excluding reporting and visualization) can be interconnected, and reporting and visualization can be performed subsequently for display.
% Table generated by Excel2LaTeX from sheet 'unclear taxonomy'

We introduce the involved tasks one by one as follows:

1. \textbf{Rudimentary Operations}: These operations encompass a set of functions and procedures that can be executed using the Structured Query Language (SQL)~\cite{date1989guide}. Their primary purpose is to enable users to perform data management, manipulation, and transformation on multi-dimensional structured data. The main operations include group by, aggregation, filter, sort, and so on. 

2. \textbf{Basic Insights}: In the context of a multi-dimensional dataset, an insight represents an interesting observation about a particular subject from a specific perspective~\cite{quickinsight, ma2021metainsight, chen2009toward}. Text2Analysis incorporates seven commonly insights:

\begin{itemize}[leftmargin=0pt,itemindent=\parindent]
    \item Rank: Within a group comprising multiple values, the highest value significantly exceeds all other values.
    \item RankLast: Within a group comprising multiple values, the lowest value is notably smaller than all other values.  
    \item Attribution: In a group of multiple non-negative values, the highest value is equal to or larger than the sum of all other values.  
    \item Trend: A time series (segment) exhibits an increasing or decreasing trend.  
    \item Monotonicity: A time series (segment of) exhibits a consistent and unidirectional increasing or decreasing trend. 
    \item Outlier: A time series contains outliers, which deviate significantly from the trend compared to the majority of points and their neighbors.
    \item Unimodality: A (segment of) time series  exhibits an unimodal distribution, characterized by a single peak or turning point, and may display U-shaped patterns.
\end{itemize}

% 3. \textbf{Forecasting}: Forecasting is the task of predicting future trends, behaviors, or outcomes based on the analysis of historical data. %\TODO{Xinrun polish}
3. \textbf{Forecasting}: Forecasting involves predicting future trends and outcomes by analyzing historical data using statistical methods, machine learning algorithms, and time series models \cite{taylor2018forecasting,hosseini2021flexible}. This process identifies patterns and relationships within the data, enabling informed predictions about future events.

4. \textbf{Chart Generation}: Chart generation refers to the recommendation and construction of prevalent charts derived from a given table~\cite{moritz2019formalizing, luo2018deepeye, zhou2021charts}.

We choose commonly used Python libraries for each task as follows, to address the corresponding analysis query:
\begin{itemize}[leftmargin=0pt,itemindent=\parindent]
    \item Rudimentary Operations: Pandas\footnote{https://pandas.pydata.org/} (APIs excluding \textit{plotting}\footnote{https://pandas.pydata.org/docs/reference/plotting.html}).
    \item Each task of Basic Insights: Custom functions are implemented to perform the mentioned tasks, and provide results for evaluation. 
    % The definition of custom functions are shown in \refsec{sec:app_task}.
    \item Forecasting: Greykite\footnote{https://github.com/linkedin/greykite} (\textit{Forecaster}), Prophet\footnote{https://github.com/facebook/prophet} (\textit{Prophet}).
    \item Chart Generation: Matplotlib\footnote{https://matplotlib.org/} (\textit{pyplot}).
\end{itemize}

\subsection{Unclear Queries}
\label{sec:unclear_tax}

\begin{table*}[htb]
  \centering
  \small
  
    \begin{tabular}{ccl}
    \toprule
    \textbf{Tasks} & \textbf{Parameters} & \textbf{Meanings of Parameters and Missing Parameters Query} \\
    \midrule
    \multicolumn{1}{c}{\multirow{5}[2]{*}{Rudimentary Operations}} & clear & E.g., Which brand has the highest total sales in 2023? \\
    \cmidrule(l){2-3}
          & field (msr\_field) & 
          \begin{tabular}[c]{@{}l@{}}Measure field for sort or aggregation.\\      E.g., Which brand had the best overall in 2023?\end{tabular}
           \\
    \cmidrule(l){2-3}
          & agg (agg\_func)    & \begin{tabular}[c]{@{}l@{}}Aggregation function, such as   sum, average…\\      E.g., Which brand has the highest sales in 2023?\end{tabular}\\
    \midrule
    \multirow{3}[2]{*}{Basic Insights } & clear & E.g., Does total increase over time? \\
    \cmidrule(l){2-3}
    %       & msr   & \begin{tabular}[c]{@{}l@{}}Measure field   for the insight.\\      E.g., Is there an increase over time?\end{tabular} \\
    % \cmidrule(l){2-3}
    %       & dim   & \begin{tabular}[c]{@{}l@{}}Dimension field for the   insight.\\      E.g., Does total increase?\end{tabular} \\
          & field   & \begin{tabular}[c]{@{}l@{}}Field for the   insight.\\      E.g., Is there an increase over time?\end{tabular} \\
    \midrule
    \multirow{5}[2]{*}{Forecasting} & clear & E.g., Forecast the cost data of different brands, categories and models of cars in 2012. \\
    \cmidrule(l){2-3}
          & forecast field & 
          \begin{tabular}[c]{@{}l@{}}Measure field   used for forecasting.\\      E.g., What will be for different categories and models of cars in 2012? \end{tabular}  \\
    \cmidrule(l){2-3}
          & steps / freq & \begin{tabular}[c]{@{}l@{}}Forecasting steps and/or   frequency.\\      E.g., What will be the sales and cost data of different brands, categories and models of cars? \end{tabular} \\
    \midrule
    \multirow{7}[2]{*}{Visualization} & clear & E.g., Help me create a bar chart to visualize the Frequency field for the HH field. \\
    \cmidrule(l){2-3}
          & chart type & 
          \begin{tabular}[c]{@{}l@{}}Char type,   including lineChart, barChart, scatterChart, pieChart.\\      E.g., Help me create a chart to visualize the Frequency field for the HH field. \end{tabular} \\
    \cmidrule(l){2-3}
          & x fields & 
          \begin{tabular}[c]{@{}l@{}}Fields for x-axis.\\      E.g., Help me create a bar chart to visualize the Frequency field.\end{tabular}\\
    \cmidrule(l){2-3}
          & y fields & \begin{tabular}[c]{@{}l@{}}Fields for y-axis.\\      E.g., Help me create a bar chart to visualize for the HH field.\end{tabular}
           \\
    \bottomrule
    \end{tabular}%
  \caption{Taxonomy and Examples of Unclear Queries}
  \label{tab:unclear_para}%
  \vspace{-2mm}
\end{table*}%

% There are two reasons for the emergence of unclear queries. Firstly, 
In many real situations, users do not directly provide complete queries, but rather give queries with some unclear intents. There are various ways to address them, such as recommending completions for the missing intents or guiding users to complete the query. This paper focuses on proposing a benchmark and does not explore the solution methods in depth. We only use the model for recommendations, which can also satisfy the exploration of the next purpose.

Secondly, the analysis and recommendation capabilities of large language models can be explored through unclear queries. When recommending for unclear queries, the model not only needs to possess semantic parsing capabilities but also requires analytical recommendation capabilities. Exploring these capabilities of large language models is crucial for better utilizing them in the analytical domain.

An unclear query lacks the essential information required to perform tasks. In other words, in the query, there are missing parameters for generating the analysis code which consists of operations from the chosen libraries.
 Since the same task may require different parameters in different libraries, we have combined the representatively used libraries for each task in \refsec{sec:ana_tax} and selected the essential parameters as shown in \reftab{tab:unclear_para}. Some parameters are not provided for missing parameters as follows:
 
\begin{itemize}[leftmargin=0pt,itemindent=\parindent]
\item When a parameter is absent, the associated operator will be excluded from use. 
% For instance, if a \textit{filter value} is unavailable, the filter operation will be deemed unnecessary. 
The parameters for this scenarios are \textit{dimension field} for rudimentary operations, \textit{filter condition} for rudimentary operations, \textit{insight type} for basic insights.
\item Parameters are typically not mentioned in the query or possess standard default values. 
% For instance, although forecasting necessitates a \textit{time field}, most queries do not explicitly specify it. Nevertheless, a time field can be chosen from the table to serve as the default value in a forecast analysis. 
The parameters for similar scenarios are \textit{confidence} for forecasting, \textit{p-value} for basic insights, \textit{measure aggregation} for basic insights.
\end{itemize}
 
 % Examples of missing parameters for each task can be found in \reftab{tab:unclear_para} and details of parameters are described in \refsec{sec:app_task}.

\begin{figure*}[htb]
    \centering
   \includegraphics[width=2 \columnwidth]{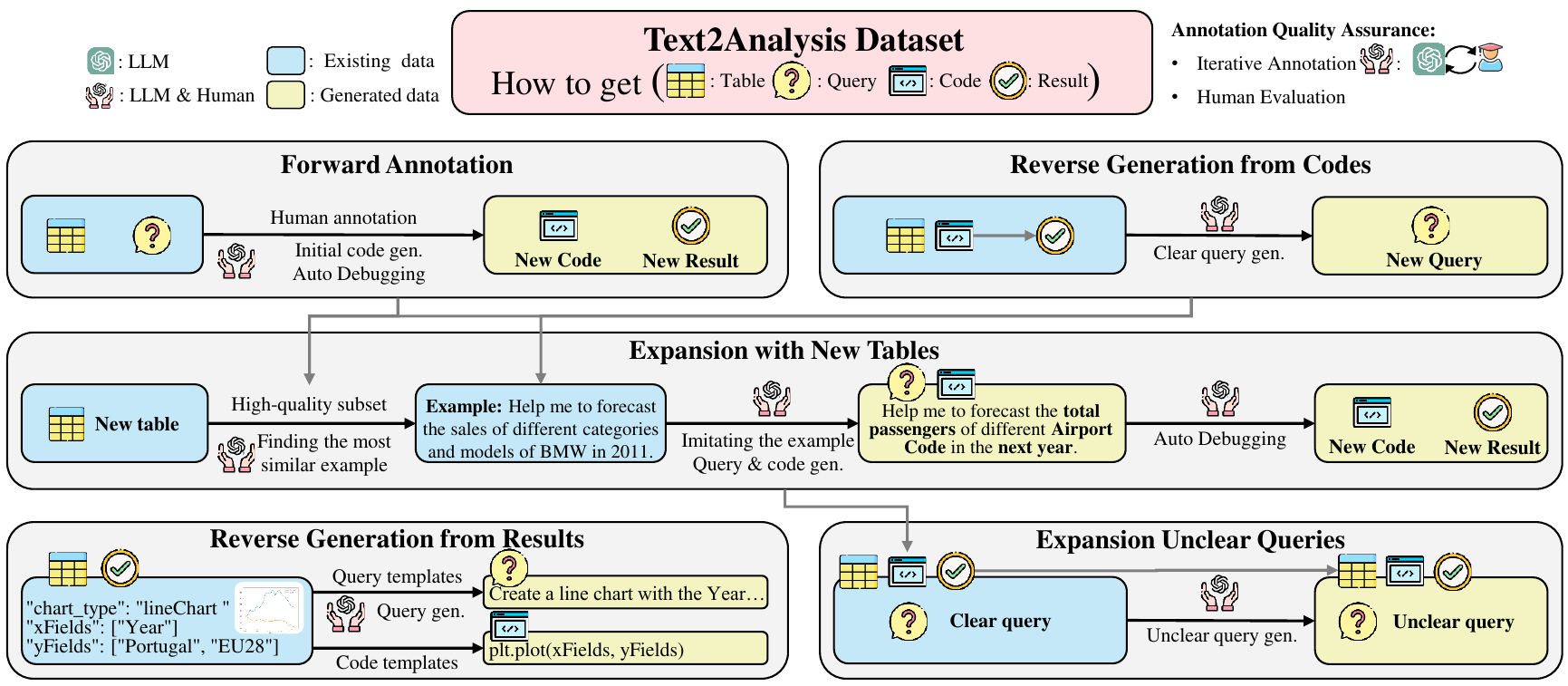}
    \caption{Collection and Generation of (table, query, code, result) Tuples in Text2Analysis.}
    \label{fig:dataset}
    \vspace{-3mm}
\end{figure*}

In addition to missing parameters, there are other types of unclear queries, such as, ambiguous parameters, unclear tasks. For ambiguous parameters, a query may have all parameters provided, but they are ambiguous or vague. \Eg, a table has two fields, UnitPrice and TotalPrice, but the query only mentions ``price'', resulting in ambiguity. Another example is when a query mentions filtering ``young people'', but there is no universally accepted definition of ``young'', leading to varied age filters. There are more details in \cite{wang-etal-2023-know}, and we will not discuss this further in this paper. For unclear tasks, a query does not explicitly specify what task to use for analysis, \eg, ``What should I do if I want to get more profits''. This query only proposes a goal without specifying any tasks, and solving such problems requires stronger problem-solving abilities. In this work, we will not discuss this further and will consider it as future work.

% According to the different ways in which parameters can be unclear, we divide them into two categories:
% \begin{enumerate}
%     \item Ambiguous parameters: The query contains terms related to the parameter, but they are not fully specified, leading to ambiguity or multiple possible interpretations. For example, 
%     \item Missing parameters: The query lacks a parameter, and there is no description of it. 
% \end{enumerate}

\section{Text2Analysis Dataset}
\label{sec:dataset}

In this section, we introduce how to collect the Text2Analysis dataset and ensure its quality. To better accomplish the following annotation, we established a tabular data annotation website based on Label Studio\footnote{https://labelstud.io/}
% and the annotation schematic is illustrated in \reffig{fig:web}.

\subsection{Data Collection}

According to the definition in \refsec{sec:problem}, our benchmark requires a quadruple $(table, query, code, result)$ for each example. However, it is extremely challenging to collect such quadruples as the annotation cost is quite high. It requires annotators with related expert backgrounds. Moreover, due to the involvement of code, the quality and time requirements for the annotators are even more stringent, resulting in an increased annotation cost. To address this issue, we propose the following ingenious and reliable annotation methods utilizing large language models as illustrated in \reffig{fig:dataset}. 

% \vspace{-1mm}
\subsubsection{Forward Annotation:} $\small(table,\ query) \rightarrow (code,\ result)$. 9 experts with extensive experience in the field of data analysis are invited to participate in the annotation process. First, we collected real tables and queries from the annotation. Subsequently, on the basis of them, the annotators labeled the code. To accelerate the annotation process, we designed an initial code generation tool and an automatic debugging tool using large language models. Annotators generated the initial version of the code, modified it, and iterated with the automatic debugging tool to ensure the accuracy and executability of the code. 

In the initial code generation tool, we used the selected libraries and encapsulated some functions for different tasks, allowing the large language model to generate code using these functions. For the automatic debugging tool, we input the original query, original code, and error messages into the large language model and carried out multiple iterations to obtain executable code.
% \vspace{-1mm}

% \subsubsection{Reverse Generation from Codes Snippets:} $\small(table,\ code,\ result) \rightarrow (query)$. We collected code and tables from major data analysis libraries and generated queries in reverse. (For Forecasting)\TODO{Xinrun}
% \vspace{-1mm}
\subsubsection{Reverse Generation from Codes Snippets:} $\small(table,\ code,\ result) \rightarrow (query)$. We have crawled and collected executable code and tables from major data analysis libraries, as detailed in Section \refsec{sec:ana_tax}. By employing a few-shot setting of GPT-4 and human verification, we reverse-engineer the data operations performed by the code to generate corresponding user-friendly natural language queries. To enhance the dataset's diversity, we adjusted parameters within the code snippets, resulting in a broader range of queries.

% \vspace{-1mm}

\subsubsection{Reverse Generation from Results:} $(table,\ result) \rightarrow (code,\ query)$. Some existing datasets provide tables and analysis result patterns but lack queries and code. We utilized these available results and designed code templates, which were then filled in to generate the corresponding code. Using the result, code, tables, and large language models, we subsequently generated queries. We perform this method for chart generation task on a high-quality subset of chart corpus in AnaMeta\cite{zhou2021charts}.

% \vspace{-1mm}
 
\subsubsection{Expansion with New Tables:} $(table) \rightarrow (query,\ code,\ result)$.  In order to enrich the diversity of tables and expand our dataset, we expand with new high-quality tables. After the previous processes, we collect a diverse sub-dataset involving various tasks. We identify high-quality tables and matched them with similar example tables in the sub-dataset. We then use large language models to mimic and generate corresponding queries, code, and results for new tables from these examples.

% \vspace{-1mm}

\subsubsection{Expansion Unclear Queries:} $(clear\ query) \rightarrow (unclear\ query)$.  In order to obtain more unclear queries, we remove the corresponding parameters from the existing queries and rewrite them. In the collected sub-dataset, we have already annotated the task taxonomy. For the corresponding task, we explicitly specify the corresponding parameter that needs to be removed for a query ($query_{old}$) and utilizes large language models to assist in generating new unclear query ($query_{new}$) in a user-oriented tone. The unclear query also requires recommended lists of corresponding codes ($code_{new}$) and results ($result_{new}$). We use the table's ($table_{old}$) existing codes ($code_{old}$) and results ($result_{old}$)  that can serve as outputs for the unclear query, forming the recommended list. Because the existing data collected for the table is filled in by real users or comes from real data, it represents the most common analysis queries and results for the table, making them suitable for forming the recommended list of unclear query outputs.

\subsection{Annotation Quality Assurance}
To ensure the quality of Text2Analysis, we performed iterative annotation and human evaluation.

% \vspace{-1mm}
\subsubsection{Iterative Annotation and Refinement}
To enhance the credibility of the large language model's generation, each time we used them for generation, we annotated and verified at least 100 samples afterward. We then modified the instructions and examples, iterating through this process for at least two rounds. This iterative approach ensured the accuracy and quality of the generated examples, resulting in a more reliable and representative benchmark dataset for evaluation purposes.
% \vspace{-1mm}

\subsubsection{Human Evaluation}

To evaluate the quality of Text2Analysis annotations, we also perform a human evaluation after the data collection is completed. We sample 100 $(table,\ query,\ code,\ result)$ pairs and invited  7 experts with data analysis experience to perform the annotation. The evaluation is conducted from five perspectives: query, task taxonomy, unclear query taxonomy, code, and result correctness using scores 0 or 1 (incorrect or correct). Each sample is annotated twice. The average score of each perspective is 0.96, 0.92, 0.93, 0.89, and 0.88, respectively. The overall average score is 0.91 and Cohen's Kappa~\cite{kappa} is 0.79 (value range:$[-1,\ 1]$). This indicates that Text2Analysis has a high annotation quality and inter-annotator agreement.

% \begin{table}[!ht]
% \small
%     \centering
%     \caption{Human Evaluation Result. ``Taxo.'' means taxonomy.}
%     \begin{tabular}{lcccc}
%     \toprule
%         Correctness & \%S $\geq$ 4 & \%S $\geq$ 3 & Agree & Kappa \\ 
%         \midrule
%         Query & ~ & ~ & ~ &    \\ 
%         Task Taxo. & ~ & ~ & ~ &   \\ 
%         Unclearness Taxo. & ~ & ~ & ~ &   \\ 
%         Code & ~ & ~ & ~ &    \\ 
%         Result & ~ & ~ & ~ &  \\
%         \bottomrule
%     \end{tabular}
%     \label{tab:anno}
% \end{table}

\subsection{Data Statistics and Distribution}
\label{sec:statis}

\begin{figure}[htbp]
    \centering
    \includegraphics[width=1\linewidth]{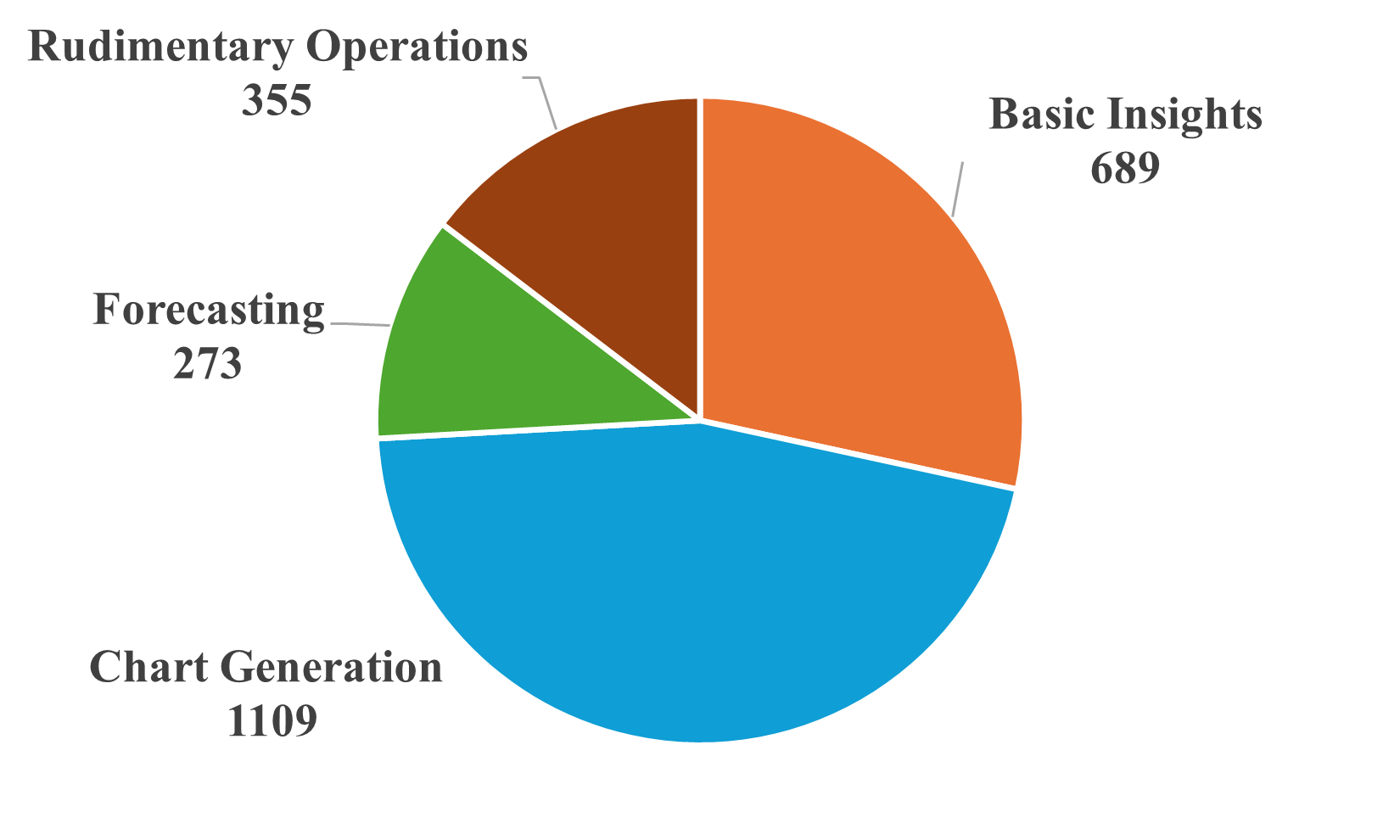}
    \caption{Analysis Task Distribution of All Queries.}
    \label{fig:sta_task}
    \vspace{-3mm}
\end{figure}

\begin{figure}[htbp]
    \centering
    \includegraphics[width=1\linewidth]{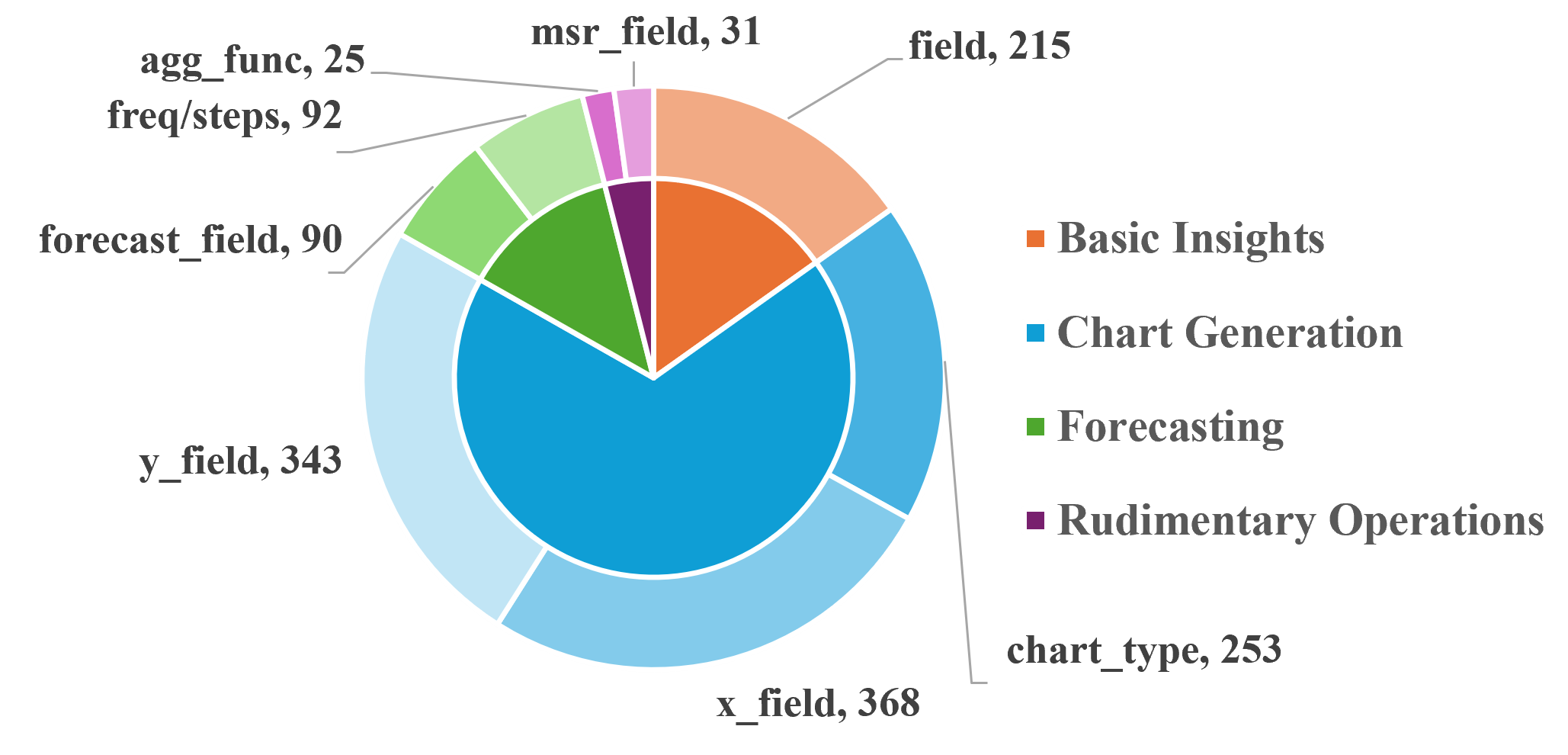}
    \caption{Task \& Parameter Distribution of Unclear Queries.}
    \label{fig:sta_unclear}
    \vspace{-3mm}
\end{figure}

Text2Analysis encompasses a total of 2249 $(table,\ query,\ code,\ result)$ pairs, sourced from 347 distinct tables. 
Queries of Text2Analysis encompass a variety of tasks, as demonstrated in \reffig{fig:sta_task}. And they encompass a diversity of unclear queries, as demonstrated in \reffig{fig:sta_unclear}. Those figures highlight the distribution of queries and code and further showcase the  diversity of the dataset and the difficulty of the problem.

\section{Evaluation Methodology}
\subsection{Baselines}
\label{sec:baseline}
We evaluate the performance of three types models namely GPT family models, code Generation models and tabular models on Text2Analysis:

\textbf{GPT family models}: GPT-4
% /GPT-3.5/ChatGPT 
models~\cite{openai2023gpt4} are potent large-scale language models with the ability to generate human-like text and high-quality code. They perform a wide range NLP tasks well with zero or few shots.
   
\textbf{Code Generation models}: 
% StarChat-$\alpha$/StarChat-$\beta$~\cite{Tunstall2023starchat-alpha} are language models specifically designed to serve as effective coding assistants, providing valuable support to programmers.  They are fine-tuned versions of the StarCoder\cite{li2023starcoder} family, which are 15.5B parameter models trained on 80+ programming languages. StarCoder family models are State-of-the-Art LLM for Code. 
% % While they are not  instruction models, which is not suitable for problems with queries. 
% StarCoder are code completion models and are not designed for instruction purposes, while Text2Analysis needs instructions about queries, tables, and task explanation. 
% Thus we use their prompted version models -- StarChat for Text2Analysis evaluation.
StarChat-$\alpha$/$\beta$~\cite{Tunstall2023starchat-alpha} and CodeGen2.5~\cite{Nijkamp2023codegen2} are language models specifically designed to serve as effective coding assistants, providing valuable support to programmers.

StarChat-$\alpha$/StarChat-$\beta$, derived from the StarCoder\cite{li2023starcoder}  family, are fine-tuned language models with 15.5 billion parameters, adept at aiding programmers across 80+ programming languages. Unlike original StarCoder models that focused on code completion, StarChat versions are better suited for Text2Analysis tasks that require query instructions and task explanations.

CodeGen2.5, an autoregressive language model built upon CodeGen2. The model is trained on StarCoderData for 1.4T tokens, achieving competitive results compared to StarCoderBase-15.5B with less than half the size.

% Table generated by Excel2LaTeX from sheet 'Exp'
\begin{table*}[htb]
\small
  \centering
  
    \begin{tabular}{l|cc|cc|cc|cc|cc}
    \toprule
    \multicolumn{1}{c|}{\multirow{2}[2]{*}{Model}} & \multicolumn{2}{c|}{Overall} & \multicolumn{2}{c|}{Rudimentary Operations} & \multicolumn{2}{c|}{Basic Insights} & \multicolumn{2}{c|}{Forecasting} & \multicolumn{2}{c}{Chart Generation} \\
          & ECR   & pass@1   & ECR   & pass@1   & ECR   & pass@1   & ECR   & pass@1   & ECR   & pass@1 \\
    \midrule
    GPT-4 & \textbf{69.46\%} & \textbf{41.01\%} & \textbf{86.76\%} & \textbf{56.82\%} & \textbf{71.41\%} & 7.31\% & 5.13\% & \textbf{35.71\%} & 79.62\% & \textbf{54.36\%} \\
    % ChatGPT & 10.11\% & 33.33\% & 48.42\% & 36.96\% & 32.99\% & 15.38\% & 0.00\% & 0.00\% & 5.20\% & 50.88\% \\
    % GPT-3.5 & 54.73\% & 49.94\% & 84.21\% & 33.75\% & 61.42\% & 13.22\% & 0.00\% & 0.00\% & 64.64\% & 58.08\% \\
    % \midrule
    StarChat-$\alpha$ & 39.60\% & 32.88\% & 61.08\% & 48.34\% & 18.16\% & \textbf{14.29\%} & 6.59\% & 0.00\% & 54.09\% & 32.84\% \\
    StarChat-$\beta$ & 60.06\% & 32.88\% & 46.20\% & 41.46\% & 41.22\% & 10.92\% & \textbf{15.75\%} & 2.33\% & \textbf{88.91\%} & 38.54\% \\
    CodeGen2.5 & 30.83\% & 19.39\% & 30.70\% & 36.70\% & 28.16\% & 18.56\% & 0.73\% & 0.00\% & 39.95\% & 15.58\% \\
    % \midrule
    TAPEX & -     & -     & -     & 11.55\%& -     & -     & -     & -     & -     & - \\
    \bottomrule
    \end{tabular}%
  \caption{Baseline Performance on Text2Analysis. (ECR = executable code ratio, pass@1 = pass rate).}
    \label{tab:exp_overall}%
\vspace{-3mm}
\end{table*}%

\textbf{Tabular models}: 
% DATER~\cite{ye2023large} is a table-based reasoning model that leverages large language models (LLMs) for enhanced performance, and it achieves state-of-the-art performance on TableQA and Text2SQL datasets such as WikiTQ\cite{pasupat-liang-2015-compositional}. 
TAPEX\cite{liu2022tapex} (Table Pre-training via Execution) is a straightforward yet highly effective pre-training method designed to enhance existing models with table reasoning capabilities. It achieves state-of-the-art performance on TableQA, Text2SQL and TabFact datasets such as WikiSQL\cite{zhongSeq2SQL2017}.

\subsection{Evaluation Metrics}

Due to the numerous tasks involved in the problem and the outputs consisting of both code and results, evaluating the generated solutions with appropriate metrics poses a challenge. We have selected three metrics to evaluate from different perspectives.
% executable code ratio, accuracy, and regression metrics. 
Executable code ratio evaluates the model's ability to generate executable code. Pass rate evaluates the correctness of the generated code. Regression scores measure the predicting capability of the chosen model within the generated code.

\textbf{Executable code ratio (ECR)}: It refers to the proportion of generated code that is executable. It evaluates the model's ability to generate executable code.
\begin{equation}
\small
ECR = \frac{\#(samples\_with\_executable\_code)}{\#(total\_samples)}
\end{equation}

% \vspace{-1mm}
% \textbf{Pass Rate (pass@1)}: It is a code generation testing metrics and refers to the proportion of generated code that is correct. In this work, there is only one single test case per code snippet, because queries target a specific table (test case).
\textbf{Pass Rate (pass@1)}: It is a metric for code generation testing that indicates the proportion of code generated accurately. Within this work, there exists only one singular test case for each code snippet, as queries are aimed at a specific table (test case). Thus, pass rate is equivalent to the accuracy in this work.
% Accuracy is equivalent to the pass rate with a single test case per code snippet. 
\begin{equation}
\small
% \begin{split}
% ACC &= \frac{\sum_{i=0}^{N}(pass\_rate_i)}{\#(samples\_with\_executable\_code)} \\
% &= \frac{\#(samples\_with\_correct\_code)}{\#(samples\_with\_executable\_code)}
% \end{split}
pass@1 = \frac{\#(samples\_with\_correct\_code)}{\#(samples\_with\_executable\_code)}
\end{equation}
% where, $N$: the number of $samples\_with\_executable\_code$.

When determining whether a code snippet passes, it should exactly match the results of executing the generated code with the results of executing the annotated code. To ensure a fair comparison, we standardize the output format and establish a checker to avoid misidentifying results with different formats but identical content as true negatives. 
% For example, we convert all results to a precision of two decimal places. Further details are described in \refsec{sec:app_eval}.
The standardized output format is as follows:
\begin{itemize}[leftmargin=0pt,itemindent=\parindent]
\item The result must be one variable, one of pd.DataFrame, List[int], List[float], List[str], int, float, str, dict.
\item For chart generation, assign a result dictionary: \{``x\_fields'': field\_name, ``y\_fields'': [field\_name1, field\_name2...], ``chart\_type'': chart\_type\}. Choose chart\_type from lineChart, barChart, scatterChart, pieChart.
\item For basic insights, answer the query in the same format as one or more outputs from our custom function.
\end{itemize}

The rules of the checker are as follows:
\begin{itemize}[leftmargin=0pt,itemindent=\parindent]
    \item For each value, we convert it to a string for comparison, including every value in lists, dictionaries, and DataFrames. If it is a number, convert it to a string after rounding to two decimal places. If it is a time, convert it to a string using a consistent format.
    \item For the comparison between DataFrames and other types: sometimes, a DataFrame is used to store a single value or a list. If the DataFrame contains only one value, that is, one row and one column, we extract it as a single value for subsequent comparisons. If there is only one column, we extract its values as a list for further comparisons.
    \item For the comparison between lists/dictionaries and other types: if a list or dictionary contains only one element, we extract it as a single value for subsequent comparisons.
    \item For comparing DataFrames with DataFrames: if the operations do not include sorting, it is acceptable for the order of rows in the DataFrames to be inconsistent. The header names of newly generated columns are allowed to differ. For missing steps/frequency forecasting, it is permissible for the predicted results to be a subset of the ground truth or vice versa, i.e., one table's rows can contain another table's rows.
\end{itemize}

\textbf{Regression scores}: For forecasting, it measures the predicting capability of the chosen model within the generated code  between the generated results and the ground truth. It includes CORR, RMSE, MAE, and MedAE.
% \vspace{-1mm}
% \begin{equation}
% regression\_score = \frac{\displaystyle\sum_{i=0}^{total\_samples}(regression\_score_i)}{\#(total\_samples)}
% \end{equation}
\begin{equation}
\small
regression\_score = \frac{\sum_{i=0}^{M}(regression\_score_i)}{\#(total\_samples)}
\end{equation}
where, $M$ is the number of $total\_samples$.

We choose to use regression scores for further evaluation because pass rate does not provide a comprehensive assessment of the quality of the generated forecasting code.  For example, if the library or forecasting model is altered in the generated code, achieving an exact match with the results executed from the annotated code can be challenging. Nevertheless, regression metrics can still be employed to assess the quality of the predictions with ground truth. 

In order to obtain the ground truth for forecasting, we preprocess the input table. During data collection, we truncate the original table according to the length of the time period required for the query's prediction. In other words, we extract a portion of the table corresponding to the prediction length as ground truth, while the remaining table is used as input for the Text2Analysis task. It is worth noting that when calculating the statistical values in \refsec{sec:statis}, truncated tables from the same original table are still considered as a single table.

\section{Experiments}

We conduct experiments on the five baselines introduced in \refsec{sec:baseline}. For the GPT family and code generation models, we design instruction prompts that include the HTML table, constraints on code generation libraries, requirements for result formatting, and so on. The parameter details for each model in the experiment are as follows:
\begin{itemize}[leftmargin=0pt,itemindent=\parindent]
    \item GPT-4: model: gpt-4-32k, temperature: 0, maximum\_length: 4096.
    % \item ChatGPT: model: ChatGPT-202301, temperature: 0, maximum\_length: 4096.
    % \item GPT-3.5: model: Text-DaVinci-003, temperature: 0, maximum\_length: 4096.
    \item StarChat-$\alpha$: model: starchat-alpha\footnote{https://huggingface.co/HuggingFaceH4/starchat-alpha}, temperature: 0.2, max\_new\_tokens: 1024.
    \item     StarChat-$\beta$: model: starchat-beta\footnote{https://huggingface.co/HuggingFaceH4/starchat-beta}, temperature: 0.2, max\_new\_tokens: 1024.
    \item CodeGen2.5: model: codegen25-7b-instruct\footnote{https://huggingface.co/Salesforce/codegen25-7b-instruct}, temperature: 0.2, max\_new\_tokens: 1024.
    \item TAPEX: model: tapex-large-finetuned-wtq\footnote{https://huggingface.co/microsoft/tapex-large-finetuned-wtq}.

\end{itemize}

\subsection{Main Results}

As shown in \reftab{tab:exp_overall}, overall experimental results demonstrate that GPT-4 outperforms other models. It achieves the highest ECR on the majority of tasks and the highest pass rate across all tasks. GPT-4's relatively better performance can be attributed to its code generation capabilities and context learning abilities. The former allows it to generate more accurate and executable code. The latter enables it to better understand and integrate the given query and instructions.

Code generation models exhibit overall performance that is comparable to GPT-4 in generating executable code. However, the pass rate of the generated code is relatively low, with the overall pass rate being 24.86\% lower than that of GPT-4. This can be attributed to the limited in-context learning capabilities of these models, which results in a restricted ability to capture the meaning of the given query and generate the correct code accordingly.

The tabular model is currently only capable of completing rudimentary tasks. Their performance on the Text2Analysis benchmark is subpar, with an pass rate of only 11.55\%. One reason for this is that rudimentary tasks in the benchmark involve complex pivot operations and calculations, which existing tabular models struggle with. These models excel at querying tables to find values in the original table but falter when it comes to performing complex calculations.

% Table generated by Excel2LaTeX from sheet 'Exp'
\begin{table}[htb]
\small
  \centering
    \begin{tabular}{l|cccc}
    \toprule
    \multicolumn{1}{c|}{Model} & CORR $\uparrow$  & RMSE   $\downarrow$& MAE  $\downarrow$ & MedAE $\downarrow$\\
    \midrule
    GPT4  & 0.10 & 0.27 & 0.24 & 0.27 \\
    StarChat-$\beta$ & 0.16 & 0.76 & 0.76 & 0.73 \\
    \bottomrule
    \end{tabular}%
  \caption{Regression Scores for Forecasting. We show the models with the highest ECR and pass@1 in \reftab{tab:exp_overall}. 
    $\downarrow$ means that smaller is better. 
  CORR is better when its absolute value is closer to 1. Thus when CORR $\in [0,\ 1]$, larger is better. 
  % $|r|\rightarrow1$ represents that the absolute values are better when they are closer to 1.}
  }
\label{tab:forecasting}%
  \vspace{-4mm}
\end{table}%

\reftab{tab:forecasting} presents the regression scores for the forecasting task, demonstrating that the code generated by the baselines has limited prediction capabilities. To successfully tackle forecasting task, not only do the models need to generate correct code, but they also need to select appropriate prediction models. The current baseline models fall short in these aspects, highlighting the need for further development and research to improve their performance on forecasting tasks.

The overall performance of the baseline models leaves room for improvement.
Its best performance is on rudimentary operations tasks, with an pass rate of only 57\%. Moreover, on complicated forecasting, it can only reach an pass rate of 14\%. This highlights the significant exploration space that still exists, presenting opportunities for further research and the development of more advanced models.

\subsection{Unclear Queries Results}

\begin{figure}[htb]
    \centering
    \includegraphics[width=1\linewidth]{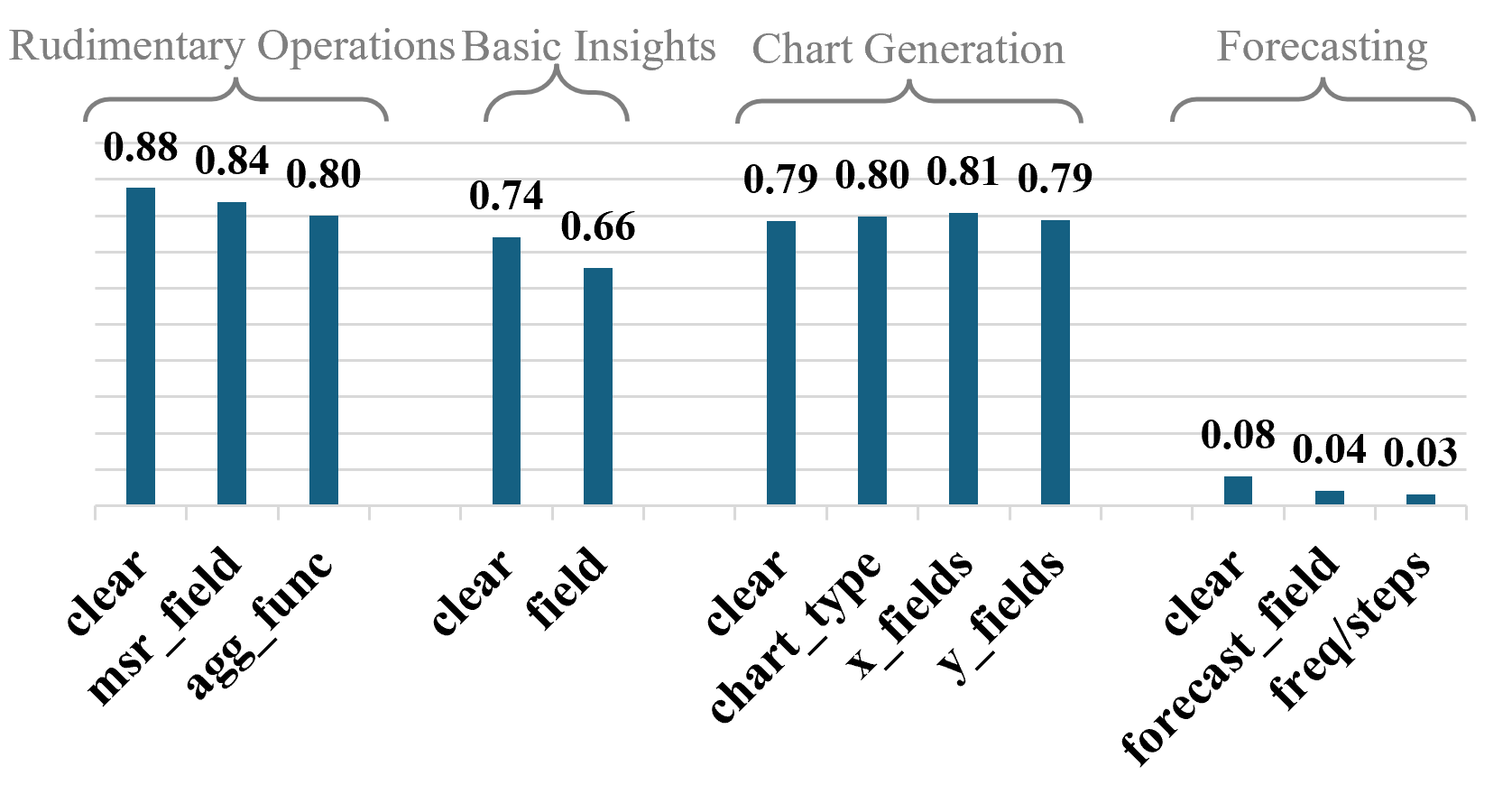}
    \caption{ECR for Unclear Queries on GPT-4.}
    \label{fig:unclear_ECR}
    \vspace{-2mm}
\end{figure}

\begin{figure}[htb]
    \centering
    \includegraphics[width=1\linewidth]{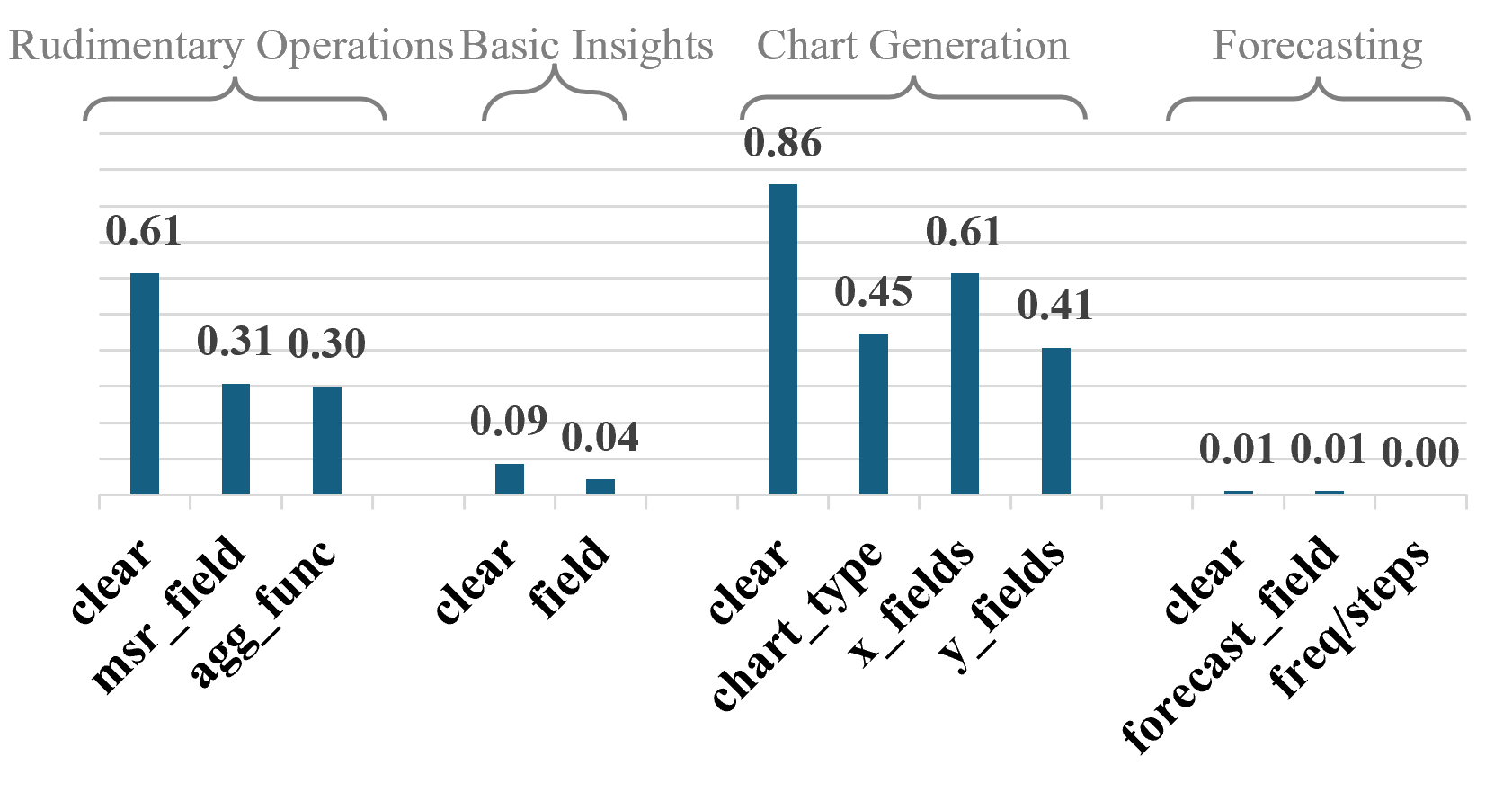}
    \caption{Pass rate for Unclear Queries on GPT-4.}
    \label{fig:unclear_acc}
    \vspace{-2mm}
\end{figure}

When facing clear queries, large models have strong parsing and code generation capabilities for data analysis. As shown in \reffig{fig:unclear_acc}, especially when the query is clear, the pass rate of chart generation is as high as 86\%. When facing a clear query, the large model needs to first parse the natural language into corresponding tasks and parameters, and then generate the correct code.

The ability to recommend fields for advanced data analysis tasks, particularly measure fields (columns with numerical attributes in a table), can be enhanced in large language models.
As shown in \reffig{fig:unclear_ECR}, the ECR decreases by 8\% on the basic insights task when the field is missing. As shown in \reffig{fig:unclear_acc}, when the measure field is missing, the pass rate has decreased for most tasks, especially the chart generation task, which has decreased by 25\%. If we want to improve the recommendation analysis, future work needs to consider injecting the knowledge of recommending analytical columns into the large language models.

The code generation capability for more complex libraries needs to be enhanced. As shown in \reffig{fig:unclear_ECR} and \reffig{fig:unclear_acc}, both the ECR and pass rate for forecasting tasks are below 1\%. The forecasting task library involves more complex operations such as parameter input and model training. In more than 50\% of the cases, GPT-4 generates incorrect parameters or input parameters that are not included in the operations. Additionally, in some instances, the code does not select the correct model, rendering it unable to successfully fit the data.

\section{Related Work}
\subsection{Tabular Benchmark}

TableQA and Text2SQL are prevalent tasks in tabular data analysis. These tasks entail answering user queries based on the information present in a source table. Notable their datasets include WikiTableQuestions ~\cite{pasupat-liang-2015-compositional},  WikiSQL~\cite{zhongSeq2SQL2017} and so on. Although numerous related datasets encompass a wide variety of table types, the primary focus remains on descriptive data analysis. Text2Analysis dataset expands to more tabular analysis tasks.

To address tabular tasks, pre-trained models like TAPAS~\cite{Herzig2020TaPasWS}, TAPEX\cite{liu2022tapex}, etc. have been employed. Concurrently, large language models have also been used in approaches like DATER~\cite{ye2023large}, StructGPT~\cite{Jiang-StructGPT-2022}, etc. In this work, we evaluate SOTA tabular models as comparison baselines.
% \vspace{-1mm}
\subsection{Large Language Models}

Recent advancements in large language models (LLMs) like GPT-3.5 and GPT-4 ~\cite{openai2023gpt4} have improved few-shot prediction capabilities and human instruction following. And they have shown promising capabilities to accelerate tabular data analysis~\cite{chen-2023-large,ye2023large, ma2023demonstration, Jiang-StructGPT-2022}. Text2Analysis is proposed as a new benchmark to further explore LLMs' upper limits in challenging tabular data analysis tasks. 

\section{Conclusion}
In conclusion, we have presented the Text2Analysis dataset that addresses the research gap in advanced analysis tasks and unclear queries in the context of tabular data analysis. Our dataset provides a comprehensive taxonomy of advanced analysis and unclear queries, which enables the evaluation of the analytical abilities of large language models. We have also proposed five innovative and reliable annotation methods that leverage large language models to accelerate the annotation process and increase the volume of annotation. Our evaluation of five SOTA models on the Text2Analysis dataset reveals their strengths and weaknesses in handling advanced analysis tasks and unclear queries, providing valuable insights for future research.

\section{Acknowledgments}

We thank all the anonymous reviewers for their valuable comments. Xinyi He and Zejian Yuan were supported in part by the National Key R\&D Program of China (2023YFB4704900) and NSFC (61976170, 62088102).

\bibliography{aaai24}

\end{document}